\documentclass[10pt,twocolumn,letterpaper]{article}

\usepackage{cvpr}      

\usepackage{hyperref}
\usepackage{graphicx}
\usepackage{amsmath}
\usepackage{booktabs}
\usepackage{multirow}

\title{ \ours: Dynamic Prompt Routing\\
    for Zero-Shot Spatial Reasoning}

\author{
Pawat Chunhachatrachai$^{1}$, Gueter Josmy Faure$^{1}$, Hung-Ting Su$^{2}$, Winston H. Hsu$^{1}$\\
$^{1}$National Taiwan University\\
$^{2}$Delta Robotics Innovation Center\\
{\tt\small pawatchun@cmlab.csie.ntu.edu.tw}
}

\newcommand{\ours}{\textsc{SpatioRoute}}
\newcommand{\oursR}{\textsc{SpatioRoute-R}}
\newcommand{\oursL}{\textsc{SpatioRoute-L}}

\begin{document}
\maketitle

\begin{abstract}
Spatial question answering over egocentric video is a challenging
task that requires Vision-Language Models~(VLMs) to reason about
3D object positions, scene affordances, and directional relationships, particularly in the
zero-shot setting where no task-specific fine-tuning is available. We introduce \ours{}, a 
dynamic prompt generation approach that routes each incoming
question to a semantically tailored prompt template---without any
additional training, fine-tuning, or 3D sensor input. \ours{}
operates in two complementary modes: \oursR{}, a rule-based router
that deterministically maps question typologies (\eg,
\textit{What}, \textit{Is}, \textit{How}, \textit{Can},
\textit{Which}) to specialized prompt templates; and \oursL{}, an
LLM-driven approach that generates task-specific prompts from the
question and situational context alone, with no video input at
routing time. We evaluate \ours{} on the SQA3D
benchmark~\cite{ma2023sqa3d} across VLMs spanning model
families. \ours{} achieves
consistent overall accuracy gains up to $5$\% over fixed prompt
baselines, establishing a new state-of-the-art for zero-shot
video-only spatial VQA without requiring 3D point-cloud inputs. As
an additional finding, we observe that Chain-of-Thought~(CoT) prompting, implemented
via the Think it Twice~\cite{tian2025thinktwice} architecture,
consistently degrades performance in this setting on Qwen series
models,
confirming that question-aware routing is more effective than
uniform reasoning instructions for spatial video understanding.
\end{abstract}
    

\begin{figure}[h]
    \centering
    \includegraphics[width=0.48\textwidth]{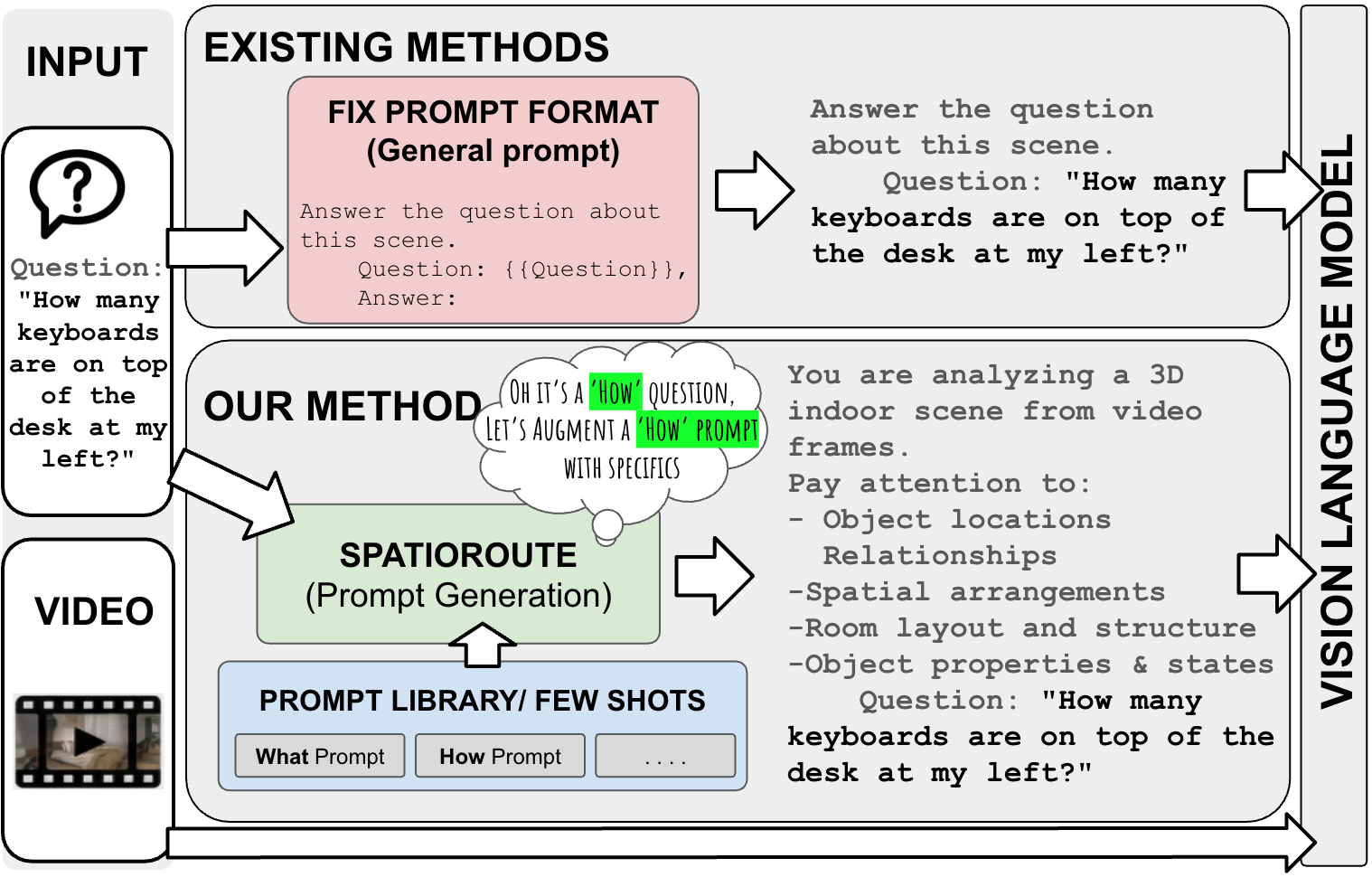}
    \caption{Comparison of existing fixed-prompt methods and \ours{}.
    \textbf{Top:} A fixed general prompt is applied uniformly
    regardless of question type, passing the raw question directly
    to the VLM. \textbf{Bottom:} \ours{} identifies the question
    type (\eg, \textit{How}) and augments it with a semantically
    appropriate prompt from the template library, providing richer
    spatial context to the VLM before answering.}
    \label{fig:overview}
\end{figure}
\section{Introduction}

\label{sec:intro}

Spatial question answering over egocentric video requires models to
reason about 3D object positions, scene affordances, and directional
relationships---a challenging setting formalized by
SQA3D~\cite{ma2023sqa3d}. While Vision-Language Models~(VLMs) such
as Qwen3-VL~\cite{bai2025qwen3vl} have shown strong general multimodal
capabilities, they remain limited in 3D spatial
understanding~\cite{chen2024spatialvlm}, motivating work on
specialized training~\cite{chen2024spatialvlm}, keyframe-based
prompting~\cite{taguchi2025spatial}, and structured reasoning
frameworks~\cite{fei2024video}.

A key challenge is that spatial VQA is fundamentally heterogeneous
in its reasoning demands. A question such as ``\textit{How many
chairs are there?}'' requires focused counting from dense visual
detail, while ``\textit{Which direction should I turn?}'' demands
egocentric directional inference, and ``\textit{Can I sit on
that?}'' requires brief affordance judgment. A single fixed prompt
cannot simultaneously serve all these divergent demands, yet
existing zero-shot approaches apply a uniform prompting strategy
regardless of question type.

We propose \ours{}, a \textbf{query-conditioned dynamic prompt
generation approach} that routes each question to the most
appropriate prompt template at inference time---requiring no
fine-tuning, no labeled data, and no 3D sensor input. \ours{}
operates in two complementary modes: \oursR{} classifies each
question by its interrogative type and deterministically maps it
to a curated prompt template; \oursL{} uses a text-only LLM with
few-shot demonstrations to generate a task-specific prompt from the
question and situational context alone, without observing the video.
Evaluated on SQA3D~\cite{ma2023sqa3d} using video-only inputs,
\ours{} achieves consistent accuracy gains of $+2$--$5$\% over
the Baseline across all Qwen model families~\cite{wang2024qwen2vl,
bai2025qwen25vl, bai2025qwen3vl}, and $+0.9$\% on
Llama3.2-11B~\cite{dubey2024llama3}, without requiring 3D
point-cloud inputs or fine-tuning.

In the course of our evaluation, we also observe that 
Chain-of-Thought~(CoT) ~\cite{wei2022chain,kojima2022large},
implemented following the Think it Twice~\cite{tian2025thinktwice}
architecture on VLMs, consistently \emph{degrades}
spatial reasoning accuracy by up to 8\%. This is consistent
with recent findings that CoT, including
Self-Consistency CoT~\cite{wang2022selfconsistency}, fails on spatial
video tasks~\cite{yang2025thinking,awal2023investigating}, and
further underscores why question-aware routing is more effective than
uniform reasoning instructions in this setting.

\noindent\textbf{Contributions:}
\begin{enumerate}[leftmargin=*, topsep=2pt, itemsep=1pt]
\item We demonstrate that \textbf{matching the prompt to the
    cognitive demand of each query} consistently outperforms both
    fixed prompts and uniform chain-of-thought reasoning across
    five VLMs---confirming that proper prompt design is a more
    effective and practical strategy than increasing model
    complexity for zero-shot spatial VQA.
   \item We introduce \ours{}, a dual-mechanism query-conditioned
dynamic prompt generation approach (\oursR{} and \oursL{}) that
achieves consistent accuracy gains of $+0.9$--$+4.7$\% over
fixed baselines across five VLMs spanning two model families,
without fine-tuning, labeled data, or 3D point-cloud inputs.

    \item We provide fine-grained per-category analysis showing that
    \oursR{} consistently improves \textit{What}, \textit{Is}, and
    \textit{Which} across all Qwen models, while \oursL{} further
    captures nuanced semantics on \textit{How} and \textit{Can}---
    demonstrating that heterogeneous prompting outperforms any
    uniform strategy.

    \item As a secondary finding, we confirm that standard CoT
    consistently degrades spatial video reasoning on Qwen models,
    while showing an opposite effect on Llama3.2-11B---
    revealing that CoT behavior is architecture-dependent and
    motivating question-aware prompting as a more robust alternative.
\end{enumerate}

\section{Related Work}
\label{sec:related}

\subsection{Spatial Reasoning in VLMs}

Improving spatial reasoning in VLMs is an active research direction.
SpatialVLM~\cite{chen2024spatialvlm} (CVPR 2024) identified a
fundamental gap due to limited 3D spatial knowledge in standard
training data, and addressed it through large-scale spatial VQA
data synthesis. SpatialRGPT~\cite{cheng2024spatialrgpt} (NeurIPS
2024) improved spatial perception through region-level grounding
and depth-aware encoding. SpatialPrompting~\cite{taguchi2025spatial}
achieved competitive zero-shot spatial reasoning via keyframe-driven
prompting without 3D inputs---the closest prior work to our setting.
VSI-Bench~\cite{yang2025thinking} (CVPR 2025 Oral) and
MSQA~\cite{yang2024msqa} (NeurIPS 2024) expanded evaluation of
video-based situated spatial reasoning, while Video-3D
LLM~\cite{zheng2025video3dllm} and zero-shot 3D QA via token
compression~\cite{huang2025zeroshot3d} (both CVPR 2025) advanced
video-only 3D scene understanding. Unlike these approaches, \ours{}
improves spatial VQA purely through query-conditioned prompt
generation, requiring no architectural changes or 3D inputs.

\noindent\textbf{Why SQA3D.} We focus exclusively on
SQA3D~\cite{ma2023sqa3d} rather than general VQA benchmarks because
it uniquely requires \emph{situated} reasoning: each question is
grounded in the observer's egocentric position and orientation within
a 3D indoor scene. The model must integrate the situational context
$s$ (``I am standing near the sofa, facing the door'') with visual
scene content to answer correctly. This egocentric grounding
distinguishes SQA3D from standard VQA benchmarks, where questions
are scene-level and observer-agnostic, making it the most
challenging and appropriate benchmark for evaluating our
query-conditioned prompting approach.

\subsection{Prompt Engineering and Adaptive Prompting}

The quality of prompts is a critical determinant of VLM performance.
Automatic Prompt Engineer~(APE)~\cite{zhou2022ape} (ICLR 2023)
showed that LLM-generated instructions can match or exceed
human-engineered prompts across 24 NLP tasks.
Guo~\etal~\cite{guo2023images} (CVPR 2023) demonstrated that
image-to-text prompt conversion enables zero-shot VQA with frozen
LLMs. Compositional CoT~\cite{gandelsmanccot2023} (CVPR 2024)
showed that structuring prompt context with scene graphs improves
compositional visual reasoning. \ours{} extends this line of work
to spatial video QA by dynamically generating question-conditioned
prompts at inference time---training-free and applicable to any
off-the-shelf VLM.

\subsection{Chain-of-Thought Prompting}

Chain-of-Thought~(CoT) prompting~\cite{wei2022chain} decomposes
problems into intermediate reasoning steps and has shown strong
gains in arithmetic and symbolic tasks. Zero-shot
CoT~\cite{kojima2022large} and Self-Consistency
CoT~\cite{wang2022selfconsistency} further extended its
applicability and reliability. However, CoT is not universally
effective: its explanations can misrepresent model
predictions~\cite{turpin2023language}, and models do not always
condition their outputs on their stated reasoning
chains~\cite{lanham2023measuring}. In visual and spatial settings,
CoT consistently causes performance drops~\cite{awal2023investigating,
chen2024enhancingrl,quan2025moreisless}, and
Yang~\etal~\cite{yang2025thinking} (CVPR 2025 Oral) found that
CoT, Self-Consistency CoT, and tree-of-thoughts all fail on spatial
video tasks. In our experiments, we reproduce this finding and show
that \ours{}'s question-aware routing overcomes this limitation.
\begin{figure*}[t]
    \centering
    \includegraphics[width=\textwidth]{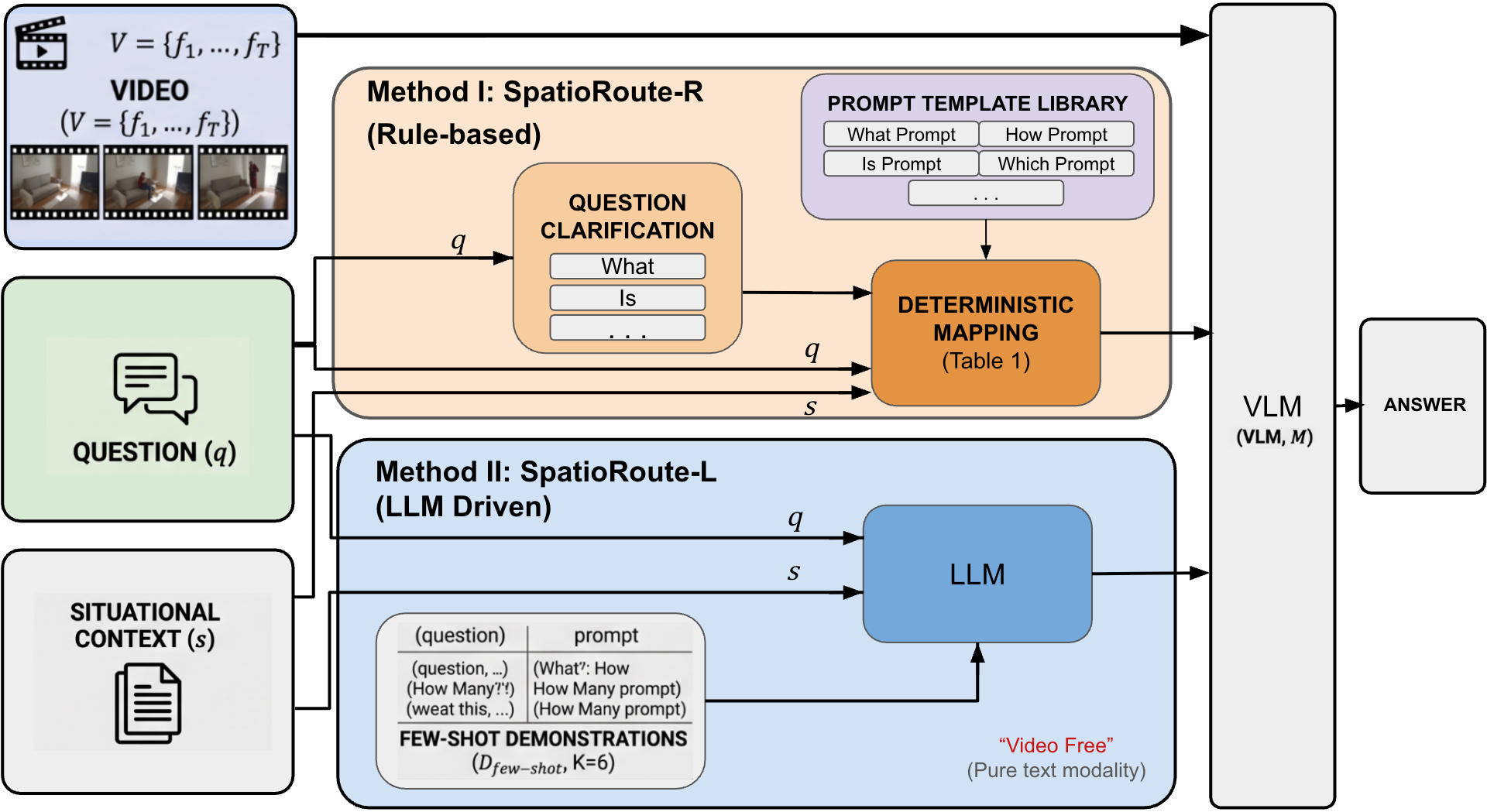}
    \caption{Overview of the \ours{}. Given a question $q$
    and situational context $s$, \textbf{Method~I: \oursR{}} (top)
    classifies the question type via lightweight string matching and
    deterministically selects a prompt template from a curated library
    \textbf{Method~II: \oursL{}} (bottom) feeds $q$ and $s$ into a
    text-only LLM conditioned on $K{=}6$ few-shot demonstrations
    $\mathcal{D}_{\text{few-shot}}$ to synthesize a task-specific
    prompt 
    ---with \emph{no video input}
    (pure text modality). Both routes produce a prompt that is
    combined with video $V$ and passed to the VLM $\mathcal{M}$ for
    final answer generation.}
    \label{fig:framework}
\end{figure*}

\section{Methodology: \ours{}}
\label{sec:method}

We present \ours{}, a query-conditioned dynamic prompt generation approach for zero-shot video spatial reasoning. The core motivation is that \emph{different spatial question types impose fundamentally different cognitive demands on a VLM, and a single fixed prompt format cannot optimally serve all of them.} \ours{} introduces a lightweight routing layer between the raw question and the VLM, selecting the most appropriate prompt before visual reasoning begins. Given a question $q$ and situational context $s$, the router produces a prompt $\mathcal{P}$:
\begin{equation}
    a = \mathcal{M}\bigl(V,\; \mathcal{P}(q, s)\bigr), \quad \mathcal{P} = \text{\ours{}}(q, s).
\label{eq:framework}
\end{equation}

\ours{} operates through two complementary mechanisms: \oursR{} (rule-based, \S\ref{sec:rulebased}) and \oursL{} (LLM-driven, \S\ref{sec:llmdriven}).


\subsection{Problem Formulation}
\label{sec:formulation}

We consider zero-shot video spatial QA on SQA3D~\cite{ma2023sqa3d}. Each instance consists of a video $V = \{f_1,\ldots,f_T\}$ of $T$ RGB frames of a 3D indoor scene, a spatial question $q$, and a situational description $s$ of the observer's position and orientation (\eg, ``I am standing near the sofa, facing the door''). The task is to produce answer $a$ using video-only inputs, \textbf{without 3D point-cloud or depth data}, in a strict zero-shot setting with no fine-tuning or weight updates. We propose replacing a fixed template $\mathcal{P}_{\text{fixed}}$ with a \emph{query-conditioned} prompt $\mathcal{P}(q, s)$ that adapts to each question's reasoning demands.

\subsection{Question Typology}
\label{sec:typology}

Spatial questions in SQA3D cluster into six interrogative categories,
each dominated by a distinct \emph{answer type} that directly
determines what cognitive demand the prompt must activate.

\noindent\textbf{\textit{What}} is the largest category and almost
entirely open-vocabulary: the model must identify objects, describe
properties, or locate items in the scene. It demands rich visual
detail rather than binary commitment.

\noindent\textbf{\textit{How}} is a pure counting category with
no binary or directional answers---the vast majority of questions
in this group take the form ``\textit{How many...?}''. It requires
focused enumeration from dense visual observation, a task easily
disrupted by step-by-step narration that distracts from counting.

\noindent\textbf{\textit{Is}} is predominantly binary (Yes/No), but
a meaningful subset also requires directional or scene-level
reasoning before committing to a judgment. This mixed nature means
it benefits from structured, situation-grounded reasoning rather
than a simple concise answer.

\noindent\textbf{\textit{Can}} is almost entirely binary
affordance judgment---whether an action is physically possible from
the observer's position. It requires scene-level situational
awareness and demands a direct short answer; verbose elaboration
actively degrades performance.

\noindent\textbf{\textit{Which}} is predominantly directional: the
model must reason about egocentric spatial relationships
(\eg, left, right, forward, back). It requires spatial inference
grounded in the observer's orientation rather than object
description.

\noindent\textbf{\textit{Others}} is a heterogeneous mix of open,
directional, and binary answers. These instruction-type queries
benefit most from a concise, task-focused prompt that avoids
over-commitment to any single reasoning mode.

These characteristics are \emph{contradictory} for prompt design:
a verbose step-by-step prompt that aids \textit{Which} inference
actively hurts \textit{Can} and \textit{How} performance by
inducing over-elaboration. This heterogeneity is the fundamental
reason why a single fixed CoT instruction degrades spatial VQA, and
directly motivates the question-type routing in \ours{}.

\subsection{Prompt Template Library}
\label{sec:templates}

We design four prompt templates, each matched to a specific reasoning demand. All use \texttt{\{question\}} and optional \texttt{\{situation\}} placeholders.





\noindent\textbf{(T1) \texttt{details\_scene}:} Instructs the model to
act as a 3D scene analysis expert, paying attention to object
locations, spatial arrangements, room layout, and object properties.
Designed for open-vocabulary questions requiring rich scene detail.
\begin{quote}
\small\ttfamily
You are analyzing a 3D indoor scene from video frames. 
Pay attention to:\\
 - Object locations and relationships\\
 - Spatial arrangements\\
 - Room layout and structure\\
 - Object properties and states\\
 Question: \{question\} 
Provide a answer:
\end{quote}

\noindent\textbf{(T2) \texttt{step\_by\_step}:} A structured
three-step reasoning prompt incorporating situational context.
Designed for binary \textit{Is} questions requiring grounded
step-by-step inference with situational awareness.
\begin{quote}
\small\ttfamily
Analyze this scene step by step: \\
1.~Observe the spatial relationships\\
2.~Identify key elements based on described condition\\
3.~Answer based on your observation\\
Question: \{question\} Answer:
\end{quote}

\noindent\textbf{(T3) \texttt{scene\_understanding}:} A balanced,
expert-framed prompt requesting concise, observation-grounded answers.
Used for \textit{Can} binary affordance questions where brevity
is critical.
\begin{quote}
\small\ttfamily
You are an expert in 3D scene understanding. Analyze the video
frames carefully and assess whether the described action is physically
possible.
Question: \{question\} Provide a answer 
\end{quote}

\noindent\textbf{(T4) \texttt{instruction\_focused}:} A direct,
short-answer prompt minimizing over-elaboration. Used for the
\textit{Others} category containing instruction-type and
miscellaneous queries.
\begin{quote}
\small\ttfamily
You are a scene analysis assistant. Look at the video frames
and answer the question with a short, direct answer.
Question: \{question\} Short answer:
\end{quote}

\noindent\textbf{Situational Context Integration.} 
For every instance in the SQA3D dataset, the situational description $s$ provides the observer's egocentric state. To ensure the VLM effectively grounds its reasoning within this perspective, We prepend a specialized context-setting string to the prompt. 
we concatenate the following instruction: 
\begin{quote}
\small\ttfamily
Consider your current position and orientation in the scene based on. \\
Situation: \{situation\}
\end{quote}
This explicitly signals to the model that the visual reasoning must be performed relative to the coordinates and facing-direction described in $s$, rather than from a global or third-person perspective.
\subsection{SpatioRoute-R: Rule-Based Router}
\label{sec:rulebased}

\oursR{} classifies each question by its leading interrogative
token via lightweight string matching, mapping it deterministically
to a template with \emph{zero model inference overhead}. Unlike
learned or oracle routing strategies, \oursR{} requires no
training data or ground truth labels---the routing rules are
designed based on the inherent linguistic structure of each
question category.

\noindent\textbf{Routing function.} Let $\tau(q)$ denote the
question type extracted from the leading token of $q$ (\eg,
\textit{What}, \textit{Is}, \textit{How}). The routing proceeds
in two steps:
\begin{gather}
    T = \mathcal{R}_{\text{rule}}(\tau(q))
    \label{eq:select} \\
    \mathcal{P} = T(q, s)
    \label{eq:fill}
\end{gather}
where Eq.~\ref{eq:select} selects the prompt template $T$ from
the library based solely on the question type $\tau(q)$---with no
model inference involved---and Eq.~\ref{eq:fill} instantiates the
selected template by filling in the question $q$ and situational
context $s$ as placeholders, producing the final prompt
$\mathcal{P}$ to be passed to the VLM. Note that $s$ is optional
and only used in templates that require situational grounding
(\eg, \texttt{step\_by\_step} for \textit{Is} questions).

\noindent\textbf{Routing table.} Table~\ref{tab:routing} shows
the complete mapping. We highlight three non-obvious design
decisions. (1)~Both \textit{How} and \textit{What} receive
\texttt{details\_scene}, since counting requires the same rich
observation as object identification. (2)~\textit{Is} receives
\texttt{step\_by\_step} rather than a concise template, because
binary spatial questions (\eg, ``Is the lamp left of the desk?'')
benefit from structured observation before committing to yes/no.
(3)~\textit{Can} receives \texttt{scene\_understanding} rather
than \texttt{step\_by\_step}: affordance questions require
scene-level judgment but not directional reasoning, and a shorter
prompt avoids verbose over-elaboration.

\begin{table}[t]
\centering\small
\caption{Prompt template assignment per question type in \oursR{}.
Each category is mapped to a template matching its reasoning demand.}
\label{tab:routing}
\setlength{\tabcolsep}{4pt}
\begin{tabular}{lp{3.2cm}l}
\toprule
\textbf{\shortstack{Question \\ Type}} & \textbf{Reasoning Need} & \textbf{Template} \\
\midrule
\textit{What}
  & Identify objects and describe scene details
  & {\footnotesize\texttt{details\_scene}}     \\
\textit{Is}
  & Verify spatial relationship step-by-step
  & {\footnotesize\texttt{step\_by\_step}}      \\
\textit{How}
  & Count objects from dense visual observation
  & {\footnotesize\texttt{details\_scene}}      \\
\textit{Can}
  & Judge physical possibility from current position
  & {\footnotesize\texttt{scene\_understanding}} \\
\textit{Which}
  & Infer egocentric direction and orientation
  & {\footnotesize\texttt{details\_scene}}      \\
\textit{Others}
  & Follow instruction with concise direct answer
  & {\footnotesize\texttt{instruction\_focused}} \\
\bottomrule
\end{tabular}
\end{table}

\subsection{SpatioRoute-L: LLM-Driven Router}
\label{sec:llmdriven}

While \oursR{} is efficient, it relies on coarse lexical classification and cannot adapt to subtle semantic variations within a category. \oursL{} replaces the deterministic mapping with a generative prompt-synthesis step, using an LLM as a meta-reasoner:
\begin{equation}
    \mathcal{P}_{\text{gen}} = \mathcal{R}_{\text{LLM}}\bigl(q,\; s \;\big|\; \mathcal{D}_{\text{few-shot}}\bigr),
\label{eq:llmrouter}
\end{equation}
where $\mathcal{D}_{\text{few-shot}} = \{(q_i, s_i, \mathcal{P}_i^*)\}_{i=1}^{K}$ maps example question-situation pairs to reference prompts. The router LLM is instructed to identify the reasoning type demanded (\eg, counting, directional, binary) and frame the prompt accordingly.

\noindent\textbf{Video-free routing.} $\mathcal{R}_{\text{LLM}}$ never observes the video $V$, operating purely on text ($q$, $s$). This keeps routing lightweight and fully decoupled from visual inference, motivated by the observation that question semantics alone are sufficient to determine the appropriate reasoning strategy.

\noindent\textbf{Advantages of} \oursR{} and \oursL{}. handles atypically phrased questions lacking a standard interrogative token (\eg, ``Tell me the number of chairs visible'') and can blend elements from multiple templates for complex multi-aspect queries, going beyond the coarse category boundaries of \oursR{}.

\section{Experimental Setup}
\label{sec:experiments}

\subsection{Dataset}

We evaluate on SQA3D~\cite{ma2023sqa3d}, built on 650
ScanNet~\cite{dai2017scannet} indoor scenes with 6.8K situational
descriptions and 33.4K questions. The official test split contains
\textbf{3,519 questions} across six categories: \textit{What}~(1,147),
\textit{Is}~(652), \textit{How}~(432), \textit{Can}~(338),
\textit{Which}~(351), and \textit{Others}~(599). We adopt a
\textbf{video-only} setting using only RGB frames---without 3D point
clouds or depth data---which is stricter than prior
work~\cite{ma2023sqa3d} and more representative of real-world
deployment.

\subsection{Models Under Evaluation}
 
We evaluate six open-source instruction-tuned VLMs across two families: \textbf{Qwen2-VL}~\cite{wang2024qwen2vl} (2B, 7B), \textbf{Qwen2.5-VL}~\cite{bai2025qwen25vl} (3B), \textbf{Qwen3-VL}~\cite{bai2025qwen3vl} (2B), and \textbf{Llama-3.2-Vision}~\cite{dubey2024llama3} (11B). Qwen3-VL-4B with its built-in thinking mode is included as a \textbf{reference comparison only}---to examine whether a model with an internal reasoning mechanism still benefits from external prompt routing---and is evaluated under the Baseline condition alone.
\subsection{Prompting Conditions}

We evaluate \textbf{four prompting conditions} per model, forming a
controlled ablation ladder that isolates the effect of prompt strategy
from model architecture.

\noindent\textbf{(1) Baseline.} A minimal, fixed direct prompt
applied uniformly to all questions.
This serves as the common reference and is the starting prompt
inherited by all other conditions.

\noindent\textbf{(2) CoT.} We adopt the Think it
Twice~\cite{tian2025thinktwice} architecture, which applies the
Baseline direct prompt as the initial instruction and appends a
uniform chain-of-thought reasoning
step~\cite{kojima2022large,wei2022chain} before committing to an
answer. This is the most common reasoning enhancement strategy in
practice.

\noindent\textbf{(3) \oursR{} (Ours).} Our rule-based router maps
each question to a semantically appropriate template via lightweight
string matching, with zero model inference
overhead~(\S\ref{sec:rulebased}).

\noindent\textbf{(4) \oursL{} (Ours).} Our LLM-driven router
generates a task-specific prompt from $q$ and $s$ via few-shot
demonstrations~(\S\ref{sec:llmdriven}). We use
Qwen2.5-1.5B~\cite{bai2025qwen25vl} as the routing LLM---a
lightweight text-only model that operates purely on $(q, s)$
without any video input, adding minimal computational overhead
while enabling fine-grained question-aware prompt synthesis.

\subsection{Implementation Details}

\noindent\textbf{Video preprocessing.} RGB frames are uniformly
sampled from each SQA3D egocentric scan and resized to each model's
native resolution. No depth, point-cloud, or sensor data is used.

\noindent\textbf{Inference configuration.} All models run
inference-only with no gradient computation or weight updates, using
\textbf{temperature $= 0.3$} for reproducibility. Instruction-tuned
checkpoints are used throughout.

\noindent\textbf{Situational context.} The situational description
$s$ is provided where used (\ie, \texttt{step\_by\_step} template
in \oursR{} and as router input in \oursL{}). It is withheld for
Baseline and Think it Twice CoT to match standard prompting practice and
isolate the effect of prompt design.

\noindent\textbf{\oursL{} router.} A lightweight text-only LLM
generates prompts from $K{=}6$ few-shot demonstrations---one per
question category---each pairing a representative question and
situation with a reference prompt. The router generates one prompt
per question with no search or optimization loop.

\noindent\textbf{Evaluation.} All outputs are post-processed to
extract answer tokens before exact match comparison with ground
truth, following the official SQA3D protocol~\cite{ma2023sqa3d}.

\section{Results and Analysis}
\label{sec:results}

We compare four conditions: \textbf{Baseline} (fixed minimal
prompt), \textbf{CoT} (uniform step-by-step),
\textbf{\oursR{}} (rule-based dynamic routing), and
\textbf{\oursL{}} (LLM-driven prompt generation). 
Our main finding is that
\ours{} consistently outperforms the fixed Baseline across all
Qwen model families and question categories.

\subsection{Overall Accuracy}

Table~\ref{tab:overall} summarises Overall accuracy across all
models and prompting conditions. \oursR{} achieves the highest
Overall score in every Qwen model group, with consistent gains of
$+2.0$--$+4.7$\% over the Baseline across Qwen2, Qwen2.5, and
Qwen3 families. \oursL{} follows closely, with gains of
$+1.5$--$+3.7$\%, confirming that both routing mechanisms
reliably improve over a fixed prompt.

Notably, \oursR{} with Qwen3-2B achieves \textbf{50.3\%}---surpassing
the Qwen3-4B-Thinking model (47.7\%), which is a larger model
equipped with a built-in internal reasoning mode. This result
demonstrates that \textbf{question-aware prompt routing can
outperform a model's internal thinking mechanism} for spatial VQA,
suggesting that how a question is framed externally matters more
than the model's capacity to reason internally in this setting.

Llama3.2-11B shows a contrasting pattern where Simple CoT yields
the largest gain ($+4.9$\%), while \ours{} provides smaller but
consistent improvements ($+0.9$\% for \oursR{}, $+0.7$\% for
\oursL{}). This architecture-dependent behavior is discussed in
detail in \S\ref{sec:generalization}.

\begin{table}[h]
\centering
\small
\caption{Overall accuracy (\%) on SQA3D test set. \textbf{Bold}
= best per model group. $^\dagger$Reference
result. $^\ddagger$Reference comparison only;
\ours{} not applied.}
\label{tab:overall}
\setlength{\tabcolsep}{4pt}
\begin{tabular}{llc}
\toprule
\textbf{Model} & \textbf{Method} & \textbf{Overall (\%)} \\
\midrule
\multirow{4}{*}{Qwen2-2B}
  & Baseline     & 42.0 \\
  & + CoT & 38.9 \\
  & + \oursR{}   & \textbf{46.7} \\
  & + \oursL{}   & 45.7 \\
\midrule
\multirow{5}{*}{Qwen2-7B}
  & Prior work$^\dagger$ ~\cite{liu2025vlm2} & \textit{46.5} \\
  & Baseline     & 47.3 \\
  & + CoT & 42.1 \\
  & + \oursR{}   & \textbf{49.5} \\
  & + \oursL{}   & 48.8 \\
\midrule
\multirow{4}{*}{Qwen2.5-3B}
  & Baseline     & 44.8 \\
  & + CoT & 38.5 \\
  & + \oursR{}   & \textbf{47.1} \\
  & + \oursL{}   & 46.5 \\
\midrule
\multirow{4}{*}{Qwen3-2B}
  & Baseline     & 48.3 \\
  & + CoT & 47.7 \\
  & + \oursR{}   & \textbf{50.3} \\
  & + \oursL{}   & 49.9 \\
\midrule
Qwen3-4B$^{\ddagger}$ (thinking)
  & Baseline     & 47.7 \\
\midrule
\multirow{4}{*}{Llama3.2-11B}
  & Baseline     & 23.3 \\
  & + CoT & \textbf{28.2} \\
  & + \oursR{}   & 24.2 \\
  & + \oursL{}   & 24.0 \\
\bottomrule
\end{tabular}
\end{table}

\begin{table}[h]
\centering
\small
\caption{Per-category accuracy (\%) for all conditions.
\textbf{Bold} = best per model group per category.}
\label{tab:gains}
\resizebox{\columnwidth}{!}{%
\begin{tabular}{lrrrrrr}
\toprule
\textbf{Method} & \textbf{What} & \textbf{Is}
  & \textbf{How} & \textbf{Can} & \textbf{Which} & \textbf{Others} \\
\midrule
\multicolumn{7}{l}{\textit{Qwen2-2B}} \\
\quad Base.  & 35.40 & 51.07 & 37.73 & 60.36 & 35.33 & 43.98 \\
\quad \oursR & \textbf{38.45} & \textbf{57.36} & \textbf{41.20}
             & \textbf{61.83} & \textbf{45.30} & \textbf{47.42} \\
\quad \oursL & 37.49 & 56.29 & 41.90 & 61.24 & 45.01 & 44.75 \\
\midrule
\multicolumn{7}{l}{\textit{Qwen2-7B}} \\
\quad Base.  & 41.41 & 60.28 & 44.21 & \textbf{57.40} & 42.17 & \textbf{49.09} \\
\quad \oursR & \textbf{43.07} & \textbf{61.96} & \textbf{45.37}
             & 53.25 & \textbf{51.28} & 48.22 \\
\quad \oursL & 42.02 & 61.04 & 44.68 & 54.73
             & 47.86 & 49.01 \\
\midrule
\multicolumn{7}{l}{\textit{Qwen2.5-3B}} \\
\quad Base.  & 37.58 & 54.14 & \textbf{41.90} & \textbf{55.92} & \textbf{45.30} & 46.97 \\
\quad \oursR & \textbf{43.50} & \textbf{57.67} & 36.57
             & 55.33 & 42.45 & \textbf{48.16} \\
\quad \oursL & 42.46 & 56.13 & 37.27
             & 54.73 & 43.75 & 47.85 \\
\midrule
\multicolumn{7}{l}{\textit{Qwen3-2B}} \\
\quad Base.  & 41.50 & \textbf{59.82} & 46.06 & 56.21 & 50.43 & \textbf{50.80} \\
\quad \oursR & \textbf{44.12} & 58.37 & 47.22 & 56.21
             & \textbf{54.42} & 49.59 \\
\quad \oursL & 42.02 & 58.99 & \textbf{47.51}
             & \textbf{57.10} & \textbf{54.42} & 50.25 \\
\midrule
\multicolumn{7}{l}{\textit{Llama3.2-11B}} \\
\quad Base.  & 20.40 & 23.47 & 19.68 & 17.75 & 37.32 & 23.72 \\
\quad \oursR & 21.36 & \textbf{25.61} & 20.14
             & \textbf{21.89} & \textbf{37.61} & 24.37 \\
\quad \oursL & \textbf{21.62} & 24.39 & \textbf{20.60} & 21.01
             & 37.32 & \textbf{24.58} \\
\bottomrule
\end{tabular}
}
\end{table}


\subsection{Gains Per Category and Generalization}
\label{sec:generalization}

\ours{} generalizes consistently across three Qwen generations and
two parameter scales (2B--7B), with \oursR{} achieving the best
Overall in every Qwen group. Table~\ref{tab:gains} reveals several
consistent patterns in how \ours{} improves over the Baseline.

\noindent\textbf{\textit{What} and \textit{Is} — consistent
\oursR{} gains.} \oursR{} improves \textit{What} across all Qwen
models ($+3.05$\% on Qwen2-2B, $+1.66$\% on Qwen2-7B, $+5.92$\% on Qwen2.5-3B, $+2.62$\% on Qwen3-2B), confirming that the
\texttt{details\_scene} template consistently helps open-vocabulary
scene description. Similarly, \oursR{} improves \textit{Is} across
all Qwen models, with the largest gain on Qwen2-2B ($+6.29$\%),
supporting the design choice of \texttt{step\_by\_step} for binary
spatial judgment.

\noindent\textbf{\textit{Which} — strongest directional gains.}
\oursR{} delivers the most striking gains on \textit{Which} across
all Qwen models ($+9.97$\% on Qwen2-2B, $+9.11$\% on Qwen2-7B),
confirming that egocentric directional questions benefit most from
a dedicated spatial reasoning prompt. This is the category where
the gap between \oursR{} and the Baseline is most consistent and
largest in absolute terms.

\noindent\textbf{\textit{Can} — category-dependent pattern.}
\textit{Can} shows a more nuanced pattern: \oursR{} improves on
Qwen2-2B ($+1.47$\%) but the Baseline retains the best score on
Qwen2-7B (57.40\%) and Qwen2.5-3B (55.92\%). This suggests that
larger Qwen models already handle binary affordance judgment
effectively with a minimal prompt, and the
\texttt{scene\_understanding} template provides diminishing returns
as model capability increases. Notably, \oursL{} recovers some of
this gap on Qwen2-7B ($54.73$\% vs \oursR{}'s $53.25$\%), suggesting
that generative routing better captures the situational nuance
in affordance questions for stronger models.

\noindent\textbf{\textit{How} — mixed results.} \oursR{} improves
\textit{How} for Qwen2-2B ($+3.47$\%) and Qwen2-7B ($+1.16$\%),
but the Baseline retains the best score for Qwen2.5-3B (41.90\%).
\oursL{} consistently provides the best or near-best \textit{How}
score across all models, suggesting that generative routing
captures counting-specific nuances better than the fixed
\texttt{details\_scene} template.

\noindent\textbf{\oursL{} vs \oursR{}.} Across all Qwen models,
\oursL{} tends to outperform \oursR{} on \textit{How} and
\textit{Can}---the two categories requiring the most situational
context---while \oursR{} leads on \textit{What}, \textit{Is}, and
\textit{Which}. This pattern suggests that \textbf{lexical routing
suffices for structure-driven categories, while generative routing
adds value where situational context is critical.} However, \oursL{}
does not consistently outperform \oursR{} overall. We attribute
this to a \textbf{prompt over-specification problem}: when the
LLM-driven router generates a highly specific prompt tailored to
the perceived question intent, it can inadvertently constrain the
VLM's output space---directing the model toward a narrow reasoning
path that excludes valid answers. For example, a generated prompt
that instructs the model to ``focus only on objects to the left''
may cause it to ignore relevant scene context needed for a complete
answer. In contrast, \oursR{}'s fixed templates are deliberately
broader, leaving the VLM sufficient freedom to reason across the
full scene. This trade-off suggests that \textbf{the optimal
routing strategy balances specificity with generality}---precise
enough to activate the right reasoning mode, but not so narrow as
to limit the answer space.

\noindent\textbf{Llama3.2-11B.} \ours{} provides consistent gains
over the Baseline across all categories except \textit{Is} and
\textit{Mac.}, where \oursL{} slightly underperforms. \oursR{}
achieves the best score on \textit{What}, \textit{How},
\textit{Can}, and \textit{Which}, while \oursL{} wins on
\textit{Mac.} (24.58\%). The overall gains ($+0.9$\% for \oursR{},
$+0.7$\% for \oursL{}) are smaller in absolute terms than for
Qwen models, consistent with Llama's substantially lower Baseline
(23.3\% vs 42--48\% for Qwen), indicating that \ours{}'s benefit
is most pronounced in models with stronger spatial grounding
baselines.

\subsection{Additional Finding: Why CoT Fails on Qwen}
\label{sec:cotfail}

As a secondary finding, CoT consistently \emph{degrades}
performance across nearly all Qwen models and categories.
Table~\ref{tab:cot_fail} isolates the per-category accuracy change
($\Delta =$ CoT $-$ Baseline), with the heaviest damage in
\textit{Can} ($-20.12$\% , Qwen2-2B) and \textit{How}
($-18.75$\%, Qwen2-7B).


\begin{table}[h]
\centering
\small
\caption{Per-category accuracy change (\%) of CoT over
Baseline ($\Delta =$ CoT $-$ Baseline). Negative = degradation.}
\label{tab:cot_fail}
\resizebox{\columnwidth}{!}{%
\begin{tabular}{lrrrrrr}
\toprule
\textbf{Model} & \textbf{What} & \textbf{Is} & \textbf{How}
  & \textbf{Can} & \textbf{Which} & \textbf{Others.} \\
\midrule
Qwen2-2B   & $-0.96$ & $+4.14$ & $-8.10$
           & $\mathbf{-20.12}$ & $+0.28$ & $-4.95$ \\
Qwen2-7B   & $-3.92$ & $-1.08$ & $\mathbf{-18.75}$
           & $-6.81$ & $-2.57$ & $-6.62$ \\
Qwen2.5-3B & $-6.02$ & $-2.91$ & $-9.72$
           & $-11.25$ & $-5.41$ & $-7.07$ \\
Qwen3-2B   & $-2.96$ & $\pm0.00$ & $-3.24$
           & $-0.29$ & $+3.42$ & $-0.61$ \\
Llama3.2   & $+2.09$ & $\mathbf{+15.95}$ & $+1.15$
           & $+7.99$ & $-1.14$ & $+5.21$ \\
\bottomrule
\end{tabular}%
}
\end{table}
\begin{table*}[t]
\centering
\small
\caption{Comparison of methods on SQA3D. PC = point cloud,
V = video, I = images, FT = fine-tuned, ZS = zero-shot.
Methods are \emph{not directly comparable} across groups due to
fundamentally different input and training settings.}
\label{tab:sqa3d_compare}
\setlength{\tabcolsep}{5pt}
\begin{tabular}{lcccc}
\toprule
\textbf{Method} & \textbf{Input} & \textbf{Train} &
  \textbf{Extra Pipeline} & \textbf{Acc. (\%)} \\
\midrule
\textit{3D Point Cloud + Fine-tuned} \\
\quad 3D-LLM~\cite{hong20233dllm}
  & PC+I  & FT & 3D reconstruction  & 47.2 \\
\quad LEO~\cite{huang2024leo}
  & PC+I  & FT & 3D reconstruction  & 50.0 \\
\quad Chat-Scene~\cite{huang2024chatscene}
  & PC+I  & FT & 3D reconstruction  & 54.6 \\
\quad LLaVA-3D~\cite{zhu2024llava3d}
  & PC+V  & FT & 3D reconstruction  & 55.6 \\
\quad Video-3D LLM~\cite{zheng2025video3dllm}
  & V+3D  & FT & Depth + 3D pose    & 58.6 \\
\midrule
\textit{Zero-shot, No 3D Input} \\
\quad SpatialPrompting + GPT-4o~\cite{taguchi2025spatial}
  & V+(pose) & ZS & SLAM + CLIP & 52.7 \\
\quad SpatialPrompting + Gemini-2.0~\cite{taguchi2025spatial}
  & V+(pose) & ZS & SLAM + CLIP & 48.6 \\
\midrule
\textit{Zero-shot, Video-only (Ours)} \\
\quad Qwen2-7B Baseline       & V & ZS & None         & 47.3 \\
\quad Qwen2-7B + \oursR{}     & V & ZS & Negligible         & 49.5 \\
\quad Qwen2-7B + \oursL{}     & V & ZS & Tiny LLM   & 48.8 \\
\quad Qwen3-2B + \oursR{}     & V & ZS & Negligible         & \textbf{50.3} \\
\quad Qwen3-2B + \oursL{}     & V & ZS & Tiny LLM   & 49.9 \\
\bottomrule
\end{tabular}
\end{table*}

We attribute the Qwen CoT failure to a \textbf{first-thinking
bottleneck}: in the Think it Twice~\cite{tian2025thinktwice}
architecture, the model's final answer is conditioned on its
initial reasoning chain. When the first-round reasoning is
misaligned with the question type---as commonly occurs in spatial
video tasks where question demands differ significantly across
categories---the second round is constrained to refine a flawed
premise rather than reason freely, propagating the error. The
effect is most severe for \textit{Can} (binary affordance) and
\textit{How} (counting), where a verbose first-round chain
introduces scene-level narration that biases the model away from
a concise yes/no or numeric commitment. \ours{} avoids this
entirely by conditioning the prompt on the question type
\emph{before} any reasoning begins, decoupling prompt design from
the model's internal reasoning dynamics.

\subsection{Positioning Among SQA3D Methods}
\label{sec:comparison}

Table~\ref{tab:sqa3d_compare} positions \ours{} within the broader
SQA3D landscape across three fundamentally different settings.
Fine-tuned 3D methods such as Video-3D
LLM~\cite{zheng2025video3dllm} and LLaVA-3D~\cite{zhu2024llava3d}
achieve the strongest results (55.6--58.6\%) by training directly
on SQA3D with 3D point cloud or depth inputs, representing the
upper bound under a heavily privileged setting. SpatialPrompting~\cite{taguchi2025spatial}
operates zero-shot but requires a substantial preprocessing
pipeline---depth estimation, SLAM-based camera pose extraction,
and CLIP-based keyframe selection---before querying a proprietary
backbone (GPT-4o: 52.7\%, Gemini-2.0: 48.6\%). \ours{}, by
contrast, requires \textbf{negligible extra pipeline}: only RGB video and
an off-the-shelf open-source VLM. Under this strictly simpler
setting, Qwen3-2B + \oursR{} achieves 50.3\%---surpassing
fine-tuned 3D-LLM (47.2\%), approaching fine-tuned LEO (50.0\%),
and outperforming SpatialPrompting with Gemini-2.0 (48.6\%) despite
requiring zero preprocessing overhead. This demonstrates that
\textbf{question-aware prompt design is a highly practical and
infrastructure-free strategy} for zero-shot spatial VQA.

\section{Conclusion}
\label{sec:conclusion}

We presented \ours{}, a zero-shot approach for video-based spatial QA that replaces uniform instructions with query-conditioned dynamic prompting. By matching prompt structures to specific cognitive demands, \ours{} addresses the heterogeneity of spatial reasoning more effectively than fixed templates. We operationalize this via two mechanisms: \oursR{}, a deterministic rule-based router, and \oursL{}, an LLM-driven synthesis layer that operates purely on text modality to minimize overhead.

Evaluated across five VLMs, \ours{} consistently outperforms fixed baselines by up to 5\% without fine-tuning or 3D point-cloud data. Notably, Qwen3-2B with \oursR{} achieves 50.3\%, surpassing both fine-tuned 3D-LLM and Gemini-2.0-based SpatialPrompting. Our analysis further reveals that standard Chain-of-Thought prompting often degrades spatial reasoning---particularly in counting and affordance tasks---by introducing reasoning bottlenecks. \ours{} bypasses these failures by aligning the model's objective before visual inference begins. These results suggest that query-aware prompt generation is a high-impact, infrastructure-free strategy for advancing multimodal spatial intelligence.
\section*{Acknowledgement}
This work was supported by the National Science and Technology Council (NSTC), Taiwan.

{
    \small
    \bibliographystyle{ieeenat_fullname}
    \bibliography{final_bib}
    
}


\end{document}